\documentclass[runningheads,a4paper]{llncs}

\usepackage{amssymb}
\usepackage{graphicx}

\usepackage{url}
\urldef{\mailsa}\path|{alfred.hofmann, ursula.barth, ingrid.haas, frank.holzwarth,|
\urldef{\mailsb}\path|anna.kramer, leonie.kunz, christine.reiss, nicole.sator,|
\urldef{\mailsc}\path|erika.siebert-cole, peter.strasser, lncs}@springer.com|    
\newcommand{\keywords}[1]{\par\addvspace\baselineskip
\noindent\keywordname\enspace\ignorespaces#1}

\begin{document}

\mainmatter  

\title{Controlled Natural Language Processing as Answer Set Programming: an Experiment}


\titlerunning{CNL Processing as Answer Set Programming: an Experiment}

%
%
\author{Rolf Schwitter}

\authorrunning{Rolf Schwitter}

\institute{Department of Computing\\
Macquarie University\\
Sydney NSW 2109, Australia\\
\url{Rolf.Schwitter@mq.edu.au}}

%
%

\toctitle{Controlled Natural Language Processing as Answer Set Programming: an Experiment}
\tocauthor{Rolf Schwitter}
\maketitle

\begin{abstract}
Most controlled natural languages (CNLs) are processed with the help of a pipeline architecture that 
relies on different software components. We investigate in this paper in an experimental way how well answer set 
programming (ASP) is suited as a unifying framework for parsing a CNL, deriving a formal representation 
for the resulting syntax trees, and for reasoning with that representation. We start from a list of input tokens
in ASP notation and show how this input can be transformed into a syntax tree using an ASP grammar and 
then into reified ASP rules in form of a set of facts. These facts are then processed by an ASP 
meta-interpreter that allows us to infer new knowledge.
\keywords{Answer Set Programming, Controlled Natural Language Processing, Meta-programming}
\end{abstract}

\section{Introduction}

Controlled natural languages (CNLs) are subsets of natural languages whose grammars and vocabularies 
have been restricted in order to eliminate ambiguity and complexity of natural languages for automated 
reasoning~\cite{Kuhn:13,Schwitter:10}. 
These CNLs are engineered for a specific purpose and look seemingly informal like natural languages,
but they have by design the same properties as their formal target languages. Typically, the writing process 
of a CNL is supported by an authoring tool that guides the writing of a text or a question by a feedback 
mechanism~\cite{Franconi:11,Kuhn:10,Power:12,Schwitter:03}.

Most existing CNLs are processed with the help of a pipeline architecture that relies on different software 
components for parsing and translating the CNL input into a formal representation before this
representation can be processed by an automated reasoning service~\cite{Clark:05,Fuchs:08}.
In this paper, we investigate in an experimental way whether answer set programming (ASP) can 
be used as a unifying framework for CNL processing, knowledge representation and automated reasoning. 
After a brief introduction to the ASP paradigm in Section 2, we show in Section 3 how grammar rules for 
a CNL can be written in ASP and how CNL sentences can be parsed into a syntax tree. In Section 4, we 
discuss how a formal representation can be generated for these syntax trees. In Section 5,  we illustrate 
how this representation can be used for reasoning in ASP with the help of a meta-interpreter. In Section 6, 
we summarise our findings and conclude.

\section{Answer Set Programming (ASP)}

ASP is a form of declarative programming that has its roots in logic programming, disjunctive databases 
and non-monotonic reasoning~\cite{Brewka:11,Lifschitz:08}. ASP provides an expressive formal language for 
knowledge representation and automated reasoning and is based on the answer set semantics for
logic programs~\cite{Gelfond:88,Gelfond:91}. In ASP, problems are represented in terms of finite logic theories 
and these problems are solved by reducing them to finding answer sets which declaratively describe the 
solutions to these problems. An ASP program consists of a set of rules of the form:

\small
\vspace{-0.1cm}
\begin{itemize}
\item[1.]  \texttt{ h$_{1}$ ;...; h$_{m}$  :- b$_{1}$,..., b$_{n}$, not b$_{n+1}$,..., not b$_{o}$.}
\end{itemize}
\vspace{-0.1cm}
\normalsize

where \small\texttt{h$_{i}$} \normalsize and \small\texttt{b$_{i}$} \normalsize are classical literals (\small\texttt{l$_{i}$}\normalsize). 
A classical literal \small\texttt{l} \normalsize is either an atom \small\texttt{a} \normalsize or a negated atom \small\texttt{-a}\normalsize. 
A literal of the form \small\texttt{not} \normalsize  \small\texttt{l} \normalsize is a negation as failure literal. The disjunction (\small\texttt{;}\normalsize)
is interpreted as epistemic disjunction~\cite{Gelfond:14}. The part on the left of the implication (\small\texttt{:-}\normalsize) 
is the head of the rule and the part on the right is the body of the rule. If the body is empty (\small\texttt{o}\normalsize=0), then 
we omit the symbol for the implication and end up with a {\em fact}. If the head is empty (\small\texttt{m}\normalsize=0), then 
we keep the symbol for the implication and end up with an {\em integrity constraint}.  Note that ASP distinguishes between strong
 negation (\small\texttt{-}\normalsize) and weak negation (\small\texttt{not}\normalsize); these two forms of negation build the 
prerequisites for non-monotonic reasoning~\cite{Gelfond:14}. For example, the ASP program in
(2) consists of two rules, six facts and one integrity constraint:

\small
\vspace{-0.1cm}
\begin{itemize}
\item[2.]
\begin{verbatim}
successful(X) :- student(X), work(X), not absent(X).
-work(X) :- student(X), not work(X).
student(john). work(john). student(sue). work(sue). 
student(mary_ann). absent(mary_ann). 
:- student(X), cheat(X), successful(X).
\end{verbatim}
\end{itemize}
\vspace{-0.1cm}
\normalsize

\noindent This program can be processed by an ASP tool such as {\em clingo}~\cite{Gebser:11} that computes 
the following answer set:

\small
\vspace{-0.1cm}
\begin{itemize}
\item[3.]
\begin{verbatim}
{ student(john) work(john) student(sue) work(sue) student(mary_ann) 
  absent(mary_ann) successful(sue) successful(john) -work(mary_ann) }
\end{verbatim}
\end{itemize}
\vspace{-0.1cm}
\normalsize

We call an ASP program {\em satisfiable}, if it has at least one answer set. Through inspection of the above answer set, we can immediately 
see that John and Sue are successful and that Mary Ann does not work. Note that the second rule in (2) specifies 
the closed world assumption~\cite{Reiter:78} for the literal \small\texttt{work/1}\normalsize. If we add the following
facts to our program:

\small
\vspace{-0.1cm}
\begin{itemize}
\item[4.]
\begin{verbatim}
student(ray). work(ray). cheat(ray).
\end{verbatim}
\end{itemize}
\vspace{-0.1cm}
\normalsize

\noindent
then we end up with an {\em unsatisfiable} program since the situation in (4) is excluded by the constraint in (2).

\section{Writing a CNL Grammar in ASP}

The CNL that we will use in the following discussion is similar to Processable English (PENG)~\cite{White:09} 
and to Attempto Controlled English (ACE)~\cite{Fuchs:08}, but the language is less expressive since ASP does 
not support full first-order logic (FOL). However, ASP is still expressive enough to represent function-free FOL formulas 
of the \small\texttt{$\exists^{*}\forall^{*}$} \normalsize prefix class in form of a logic program~\cite{Lierler:13}.
The following text (5) is written in CNL and expresses the same information as the ASP program in (2):

\small
\vspace{-0.1cm}
\begin{itemize}
\item[5.] Every student who works and who is not provably absent is successful.
\item[] If a student does not provably work then the student does not work.
\item[] John is a student who works.
\item[] Sue is a student and works.
\item[] Mary Ann who is a student is absent.
\item[] Exclude that a student who cheats is successful.
\end{itemize}
\vspace{-0.1cm}
\normalsize


In order to process this text in ASP, we split it into a sequence of sentences and each 
sentence into a sequence of tokens. Each token is represented in ASP as a fact (\small{\texttt{token/4}\normalsize)
with four arguments: the first argument holds the string, the second argument holds the sentence 
number, and the third and fourth argument represent the start and the end position of the string, 
for example:

\small
\vspace{-0.1cm}
\begin{itemize}
\item[6.]
\begin{verbatim}
token("Every", 1, 1, 2). token("student", 1, 2, 3).  ... 
\end{verbatim}
\end{itemize}
\vspace{-0.1cm}
\normalsize

Each string is stored as a fact (\small\texttt{lexicon/5}\normalsize) in the ASP lexicon that distinguishes between function 
words and content words. Function words (e.g., {\em and, every, who}) define the structure of the 
CNL and content words (e.g., {\em student, works, successful}) are used to express the domain knowledge. 
These lexical entries contain information about the category, the string, the base form, as well as syntactic 
and semantic constraints (\small\texttt{n} \normalsize stands for nil):

\small
\vspace{-0.1cm}
\begin{itemize}
\item[7.]
\begin{verbatim}
lexicon(cnj, "and", n, n, n). 
lexicon(det, "Every", n, sg, forall).
lexicon(rp, "who", n, n, n).
\end{verbatim}
\vspace{-0.2cm}
\begin{verbatim}  
lexicon(noun, "student", student, sg, n).
lexicon(iv, "works", work, sg, n).
lexicon(adj, "successful", successful, n, n).
\end{verbatim}
\end{itemize}
\vspace{-0.1cm}
\normalsize

We can write the grammar for the CNL directly as a set of ASP rules that generate a syntax tree bottom-up,
starting from the tokens up to the root.
Let us have a look at the grammar rules that process the first sentence in (5). This sentence is interesting, 
since it contains a coordinated relative clause that is embedded in the main sentence. The first relative clause
{\em who works} is positive, and the second relative clause {\em who is not provably absent} contains a weak 
negation. It is important to note that this form of negation can only occur in a universally quantified CNL 
sentence or in a CNL sentence that results in an integrity constraint.

The grammar rule (\small\texttt{rule/7}\normalsize) specifies in a declarative way that a sentence
(\small\texttt{s}\normalsize) starts at position \small\texttt{P1} \normalsize and ends at position
 \small\texttt{P4}\normalsize, if there is a noun phrase (\small\texttt{np}\normalsize) that starts 
at \small\texttt{P1} \normalsize and ends at \small\texttt{P2}\normalsize, followed by a verb phrase
 (\small\texttt{vp}\normalsize) that starts at \small\texttt{P2} \normalsize and 
ends at \small\texttt{P3}\normalsize, followed by a punctuation mark (\small\texttt{pm}\normalsize) 
between position \small\texttt{P3} \normalsize 
and \small\texttt{P4}\normalsize:

\small
\vspace{-0.1cm}
\begin{itemize}
\item[8.]
\begin{verbatim}
rule(s, s(T1, T2, T3), n, n, N, P1, P4)  :-
  rule(np, T1, Y, n, N, P1, P2),  
  rule(vp, T2, Y, n, N, P2, P3), 
  rule(pm, T3, n, n, N, P3, P4).
\end{verbatim}
\end{itemize}
\vspace{-0.1cm}
\normalsize

The second argument position of this rule is used to build up a syntax tree, the third
argument position is used for syntactic constraints, the fourth for semantic constraints, and 
the fifth for the sentence number. The variable \small\texttt{Y} \normalsize in (8) is used to enforce 
number agreement between the \small\texttt{np} \normalsize and the \small\texttt{vp}\normalsize.

The following grammar rule in (9) further describes the noun phrase of our example sentence. 
This noun phrase (\small\texttt{np}\normalsize) consists of a determiner (\small\texttt{det}\normalsize), 
followed by a nominal expression (\small\texttt{n1}\normalsize). The variable \small\texttt{M} \normalsize 
holds a quantifier that controls -- as we will see later -- the use of weak negation in our example sentence:

\small
\vspace{-0.1cm}
\begin{itemize}
\item[9.]
\begin{verbatim}
rule(np, np(T1, T2), Y, n, N, P1, P3) :-
  rule(det, T1, Y, M, N, P1, P2), 
  rule(n1, T2, Y, M, N, P2, P3).
\end{verbatim}
\end{itemize}
\vspace{-0.1cm}
\normalsize

The nominal expression (\small\texttt{n1}\normalsize) expands in our case into a noun
 (\small\texttt{noun}\normalsize) and a relative clause (\small\texttt{rcl}\normalsize):

\small
\vspace{-0.1cm}
\begin{itemize}
\item[10.]
\begin{verbatim}
rule(n1, n1(T1, T2), Y, M, N, P1, P3) :-
  rule(noun, T1, Y, n,  N, P1, P2), 
  rule(rcl, T2, Y, M, N, P2, P3).
\end{verbatim}
\end{itemize}
\vspace{-0.1cm}
\normalsize

The noun (\small\texttt{noun}\normalsize) is a preterminal category and processes the input token
 (\small\texttt{token/4}\normalsize) with the help of the lexical information (\small\texttt{lexicon/5}\normalsize):

\small
\vspace{-0.1cm}
\begin{itemize}
\item[11.]
\begin{verbatim}
rule(noun, noun(S), Y, n, N, P1, P2) :-
  token(S, N, P1, P2), 
  lexicon(noun, S, B, Y, n).
\end{verbatim}
\end{itemize}
\vspace{-0.1cm}
\normalsize

Note that the relative clause in our example sentence is coordinated and consists of a positive and a negative part.
The grammar rule in (12) for relative clauses (\small\texttt{rcl}\normalsize) deals with this coordinated structure. 
In contrast to the positive part, we use the variable \small\texttt{M} \normalsize in the negative part of the
 coordinated structure to enforce that this form of negation occurs under universal quantification (additional 
grammar rules exist that deal with relative clauses where the order of the positive and negative part is different):

\small
\vspace{-0.1cm}
\begin{itemize}
\item[12.]
\begin{verbatim}
rule(rcl, rcl(T1, T2, T3), Y, M, N, P1, P4) :-
  rule(rcl, T1, Y, n, N, P1, P2), 
  rule(cnj, T2, n, n, N, P2, P3),
  rule(rcl, T3, Y, M, N, P3, P4).
\end{verbatim}
\end{itemize}
\vspace{-0.1cm}
\normalsize

As the following two grammar rules in (13) illustrate, the relative clause expands in both cases 
into a relative pronoun (\small\texttt{rp}\normalsize) followed by a verb phrase (\small\texttt{vp}\normalsize): 
the first \small\texttt{vp} \normalsize occurs without a variable (\small\texttt{n}\normalsize) at 
the fourth argument position and the second \small\texttt{vp} \normalsize occurs with a variable
 (\small\texttt{M}\normalsize) that holds the quantifier:

\small
\vspace{-0.1cm}
\begin{itemize}
\item[13.]
\begin{verbatim}
rule(rcl, rcl(T1, T2), Y, n, N, P1, P3) :-
  rule(rp, T1, n, n, N, P1, P2), 
  rule(vp, T2, Y, n, N, P2, P3).
\end{verbatim}
\vspace{-0.1cm}
\begin{verbatim}  
rule(rcl, rcl(T1, T2), Y, M, N, P1, P3) :-
  rule(rp, T1, n, n, N, P1, P2), 
  rule(vp, T2, Y, M, N, P2, P3).
\end{verbatim}
\end{itemize}
\vspace{-0.1cm}
\normalsize

\noindent The first verb phrase (\small\texttt{vp}\normalsize) in (13) expands into an intransitive 
verb ({\small\texttt{iv}\normalsize):

\small
\vspace{-0.1cm}
\begin{itemize}
\item[14.]
\begin{verbatim}
rule(vp, vp(T1), Y, n, N, P1, P2) :- 
  rule(iv, T1, Y, n, N, P1, P2).
\end{verbatim}
\end{itemize}
\vspace{-0.1cm}
\normalsize

\noindent and the second verb phrase (\small\texttt{vp}\normalsize) expands 
into a copula (\small\texttt{cop}\normalsize), followed by a weak negation 
(\small\texttt{naf}\normalsize) and an adjective ({\small\texttt{adj}\normalsize):

\small
\vspace{-0.1cm}
\begin{itemize}
\item[15.]
\begin{verbatim}
rule(vp, vp(T1, T2, T3), Y, M, N, P1, P4) :-
  rule(cop, T1, Y, n, N, P1, P2), 
  rule(naf, T2, n, M, N, P2, P3),
  rule(adj, T3, n, n, N, P3, P4).
\end{verbatim}
\end{itemize}
\vspace{-0.1cm}
\normalsize

As we have seen, weak negation is expressed on the surface level of the CNL with the help of the key 
phrase {\em not provably}. The rule (\small\texttt{naf}\normalsize) in (16) processes this key 
phrase if it occurs in the scope of a universal quantifier (\small\texttt{forall}\normalsize):

\small
\vspace{-0.1cm}
\begin{itemize}
\item[16.]
\begin{verbatim}
rule(naf, naf(T1, adv("provably")), n, forall, N, P1, P3) :-
  rule(neg, T1, n, n, N, P1, P2), 
  rule(adv, adv("provably"), n, n, N, P2, P3).
\end{verbatim}
\end{itemize}
\vspace{-0.1cm}
\normalsize

We still have to deal with the verb phrase (\small\texttt{vp}\normalsize) on the sentence level 
that is part of rule (8). This verb phrase expands into a copula (\small\texttt{cop}\normalsize), 
followed by an adjective (\small\texttt{adj}\normalsize):

\small
\vspace{-0.1cm}
\begin{itemize}
\item[17.]
\begin{verbatim}
rule(vp, vp(T1, T2), Y, n, N, P1, P3) :-
  rule(cop, T1, Y, n, N, P1, P2), 
  rule(adj, T2, n, n, N, P2, P3).
\end{verbatim}
\end{itemize}
\vspace{-0.1cm}
\normalsize

In ASP, these grammar rules are processed bottom-up and during this model generation 
process the following syntax tree is produced for our example sentence:

\small
\vspace{-0.1cm}
\begin{itemize}
\item[18.]
\begin{verbatim}
s(np(det("Every"),
     n1(noun("student"),
     rcl(rcl(rp("who"),
             vp(iv("works"))),
         cnj("and"),
         rcl(rp("who"),
             vp(cop("is"),
                naf(neg("not"),
                    adv("provably")),
                adj("absent")))))),
  vp(cop("is"),
     adj("successful")),
  pm("."))
\end{verbatim}
\end{itemize}
\vspace{-0.1cm}
\normalsize

\noindent This syntax tree needs to be translated into a suitable ASP representation for 
automated reasoning as we will see in Section 4.

Before we do this, please note that it is relatively straightforward to generate look-ahead information
that informs the author about the admissible input. To generate look-ahead information, we
add one or more dummy tokens that contain a special string (\small\texttt{\$lah\$}\normalsize)
to the last input token, for example:

\small
\vspace{-0.1cm}
\begin{itemize}
\item[19.]
\begin{verbatim}
token("Every", 1, 1, 2). 
token("student", 1, 2, 3). 
token("$lah$", 1, 3, 4).
\end{verbatim}
\end{itemize}
\vspace{-0.1cm}
\normalsize

\noindent Additionally, we add for each category a lexical entry that contains this special string, for example: 

\small
\vspace{-0.1cm}
\begin{itemize}
\item[20.]
\begin{verbatim}
lexicon(iv, "$lah$", n, sg, n).
\end{verbatim}
\end{itemize}
\vspace{-0.1cm}
\normalsize

The following ASP rules in (21) are then used to collect those syntax tree fragments that span the input string
and contain look-ahead information. The last rule in (21) is a constraint and makes sure that the entire input 
string is considered:

\small
\vspace{-0.1cm}
\begin{itemize}
\item[21.]
\begin{verbatim}
lah(C, T, Y, M, N, 1, P2) :- 
  rule(C, T, Y, M, N, 1, P2),
  end_pos(P2, N).
\end{verbatim}
\begin{verbatim}  
lah :- 
  lah(C, T, Y, M, N, 1, P2),
  end_pos(P2, N).
\end{verbatim}
\begin{verbatim}  
-end_pos(P2, N1) :- 
  token("$lah$", N1, P1, P2),
  token("$lah$", N2, P3, P4),
  N1 <= N2, P2 < P4.
\end{verbatim}
\begin{verbatim}  
end_pos(P2, N) :- 
  token("$lah$", N, P1, P2),
  not -end_pos(P2, N).
\end{verbatim}
\begin{verbatim}  
:- not lah.
\end{verbatim}
\end{itemize}
\vspace{-0.1cm}
\normalsize

In our example, the addition of the first token (\small\texttt{token("\$lah\$", 1, 3, 4)}\normalsize) in (19) results 
in an unsatisfiable program. The addition of a further token (\small\texttt{token("\$lah\$", 1, 4, 5)}\normalsize) 
results in two tree fragments that span the input string. From these trees, we can extract the
relevant look-ahead information.

\section{From Syntax Trees to Reified ASP Rules}

We choose an indirect encoding for ASP rules where rules are reified as facts. This kind of encoding is necessary
since there exists no mechanism within ASP that would allow us to assert new rules. The reified rules are then further 
processed by an ASP meta-interpreter as we will see in Section 5. A reified rule consists of up to four different 
fact types: a fact  type (\small\texttt{rule/1}\normalsize) for the identification of the rule, a fact type
(\small\texttt{head/2}\normalsize) for the head of the rule, a fact type (\small\texttt{pbl/2}\normalsize) for 
positive body literals (if any), and a fact type (\small\texttt{nbl/2}\normalsize) for negative body literals (if any). 
For example, the translation of the syntax tree in (18)  will result in the following encoding:

\small
\vspace{-0.1cm}
\begin{itemize}
\item[22.]
\begin{verbatim}
rule(1). 
head(1, lit(func(successful), arg(sk(1)))). 
pbl(1, lit(func(work), arg(sk(1)))).  
pbl(1, lit(func(student), arg(sk(1)))).
nbl(1, lit(func(absent), arg(sk(1)))).
\end{verbatim}
\end{itemize}
\vspace{-0.1cm}
\normalsize

The fact type (\small\texttt{rule/1}\normalsize) stores the rule number (\small\texttt{1}\normalsize). This rule 
number occurs as first argument in the other fact types and specifies rule membership. The actual literals that 
belong to a rule are encoded with the help of the term \small\texttt{lit/2} \normalsize where the first argument
 (e.g., \small\texttt{func(successful)}\normalsize) is the functor name of the literal and the second argument
(\small\texttt{arg(sk(1))}\normalsize) represents a Skolem constant\footnote{The number \textit{i} in
\textit{sk(i)}, represents the \textit{i}th Skolem constant.} that replaces the variable in the literal. Note that 
all facts that have been derived from rules need to be grounded and cannot contain any 
variables. 

In the following, we will show in detail how the syntax tree in (18) is translated into the proposed ASP notation 
for rules in (22). The syntax tree is first split into three main parts: a part (\small\texttt{to\_qnt/3}\normalsize) to be 
translated into a quantifier, a part (\small\texttt{to\_body/3}\normalsize) to be translated into a rule body, and a part
(\small\texttt{to\_head/3}\normalsize) to be translated into a rule head. In our case, these parts correspond to the 
determiner (\small\texttt{det}\normalsize),  the nominal expression (\small\texttt{n1}\normalsize), and the verb
phrase (\small\texttt{vp}\normalsize) of the first sentence in (5):

\small
\vspace{-0.1cm}
\begin{itemize}
\item[23.]
\begin{verbatim}
to_qnt(N, det("Every"), M) :-
  rule(det, det("Every"), Y, M, N, P1, P2),
  rule(n1, T2, Y, M, N, P2, P3), 
  rule(vp, T3, Y, n, N, P3, P4).
\end{verbatim}
\begin{verbatim}  
to_body(N, T2, M) :-
  rule(det, det("Every"), Y, M, N, P1, P2), 
  rule(n1, T2, Y, M, N, P2, P3), 
  rule(vp, T3, Y, n, N, P3, P4).
\end{verbatim}
\begin{verbatim}  
to_head(N, T3, M) :-
  rule(det, det("Every"), Y, M, N, P1, P2),
  rule(n1, T2, Y, M, N, P2, P3), 
  rule(vp, T3, Y, n, N, P3, P4).
\end{verbatim}
\end{itemize}
\vspace{-0.1cm}
\normalsize

In the next step, the determiner (\small\texttt{"Every"}\normalsize) is processed and this results in a new
predicate (\small\texttt{qnt/4}\normalsize) that stores a rule number (\small\texttt{R}\normalsize), the universal 
quantifier (\small\texttt{M}\normalsize), the sentence number (\small\texttt{N}\normalsize), and a Skolem constant
(\small\texttt{K}\normalsize) for the universal quantifier. Note that the rule number and the number for the Skolem 
constant are generated with the help of Lua\footnote{Lua (http://www.lua.org) is available as integrated
scripting language in {\em clingo}.} and assigned (\small\texttt{:=}\normalsize) to the variables
 (\small\texttt{R}\normalsize) and  (\small\texttt{K}\normalsize):

\small
\vspace{-0.1cm}
\begin{itemize}
\item[24.]
\begin{verbatim}
qnt(R, M, N, sk(K)) :- 
  to_qnt(N, det("Every"), M), 
  R := @rule_num(), 
  K := @sk_num().
\end{verbatim}
\end{itemize}
\vspace{-0.1cm}
\normalsize

Given the new predicate (\small\texttt{qnt/4}\normalsize) for the universal quantifier, the syntax tree fragments
for constructing the head of a rule and the body of a rule can be further split up:

\small
\vspace{-0.1cm}
\begin{itemize}
\item[25.]
\begin{verbatim}
to_head(R, adj(S2), K)  :- 
  to_head(N, vp(cop(S1), adj(S2)), M), 
  qnt(R, forall, N, K).
\end{verbatim}
\begin{verbatim}  
to_body(R, noun(S), K)  :- 
  to_body(N, n1(noun(S), RCL), M),
  qnt(R, forall, N, K).
\end{verbatim}
\begin{verbatim}  
to_body(R, RCL, K)  :- 
  to_body(N, n1(noun(S), RCL), M), 
  qnt(R, forall, N, K).
\end{verbatim}
\end{itemize}
\vspace{-0.1cm}
\normalsize

In the case of the head (\small\texttt{to\_head/3}\normalsize), this process results in a preterminal 
category (\small\texttt{adj(S2)}\normalsize) that can be used to generate the head literal of 
the rule. In the case of the body (\small\texttt{to\_body/3}\normalsize), only the first rule generates 
a preterminal category (\small\texttt{noun(S)}\normalsize) that can directly be used to generate a 
positive body literal. The second rule is used to split the relative clause ({\small\texttt{RCL}\normalsize)
into its basic constituents in order to extract the relevant preterminal categories:

\small
\vspace{-0.1cm}
\begin{itemize}
\item[26.]
\begin{verbatim}
to_body(R, RCL1, K)  :- 
  to_body(R, rcl(RCL1, cnj(and), RCL2), K).
\end{verbatim}
\begin{verbatim}  
to_body(R, RCL2, K) :- 
  to_body(R, rcl(RCL1, cnj(and), RCL2), K).
\end{verbatim}
\begin{verbatim}  
to_body(R, iv(S2), K) :- 
  to_body(R, rcl(rp(S1), vp(iv(S2))), K).
\end{verbatim}
\begin{verbatim}  
to_body(R, naf(adj(S3)), K)  :- 
  to_body(R, rcl(rp(S1), vp(cop(S2), naf(T1, T2), adj(S3))), K).
\end{verbatim}
\end{itemize}
\vspace{-0.1cm}
\normalsize

The preterminal categories for content words together with the Skolem constant (\small\texttt{K}\normalsize) 
and the rule number (\small\texttt{R}\normalsize) are then used to generate the head literal, the positive and 
negative body literals. During this process the string (\small\texttt{S}\normalsize) of these preterminal categories
 is replaced by the base form (\small\texttt{B}\normalsize) via a lexicon lookup. The rule identifier (\small\texttt{rule/1}\normalsize) 
is generated with the help of the head literal (\small\texttt{head/2}\normalsize):

\small
\vspace{-0.1cm}
\begin{itemize}
\item[27.]
\begin{verbatim}
rule(R) :- head(R, L).
\end{verbatim}
\begin{verbatim}  
head(R, lit(func(B), arg(K))) :- 
  to_head(R, adj(B), K), 
  lexicon(adj, S, B, _, _).
\end{verbatim}
\begin{verbatim}  
pbl(R, lit(func(B), arg(K))) :- 
  to_body(R, noun(S), K), 
  lexicon(noun, S, B, _, _).
\end{verbatim}
\begin{verbatim}  
pbl(R, lit(func(B), arg(K))) :- 
  to_body(R, iv(S), K),  
  lexicon(iv, S, B, _, _).
\end{verbatim}
\begin{verbatim}  
nbl(R, lit(func(B), arg(K))) :-  
  to_body(R, naf(adj(S)), K),
  lexicon(adj, S, B, _, _).
\end{verbatim}
\end{itemize}
\vspace{-0.1cm}
\normalsize

Note that checking for anaphoric references can be done over the existing model during
the translation process of the syntax tree into rules. For example, the second sentence of (5) 
contains a definite noun phrase ({\em the student}) that is anaphorically linked to an indefinite
noun phrase ({\em a student}). Depending on the context in which an anaphoric expression
occurs, we either check the body of the current rule for an antecedent or the heads of
all existing rules that don't have a body and give preference to the closest match in terms
of rule numbers.

\section{Reasoning with Reified ASP Rules}

In order to process these reified ASP rules, we use a meta-interpreter that is based on the work
of Eiter et al.~\cite{Eiter:02}. We substantially extended this meta-interpreter so that it can 
deal with variables that occur as Skolem constants in the reified notation. On the meta-level 
we represent answer sets with the help of the predicate \small\texttt{in\_AS/1} \normalsize 
and use the following two rules to add literals to an answer set: 

\small
\vspace{-0.1cm}
\begin{itemize}
\item[28.]
\begin{verbatim}
in_AS(lit(F, A2)) :-   
  head(R, lit(F, A1)), 
  pos_body_true(R, A1, A2),
  not neg_body_false(R, A1, A2).
\end{verbatim}
\begin{verbatim}  
in_AS(lit(F, A)) :-   
  head(R, lit(F, A)), 
  rule(R), 
  not pos_body_exists(R).
\end{verbatim}
\end{itemize}
\vspace{-0.1cm}
\normalsize

The first rule specifies that a literal (\small\texttt{lit/2}\normalsize) is in an answer set (\small\texttt{in\_AS/1}\normalsize), 
if it occurs in the head (\small\texttt{head/2}\normalsize) of a rule with number \small\texttt{R} \normalsize whose positive 
body (\small\texttt{pos\_body\_true/3}\normalsize) is true and whose negative body ({\small\texttt{neg\_body\_false/3}\normalsize) 
is not false and if the Skolem constant that occurs as argument (\small\texttt{A1}\normalsize) of that literal can be replaced 
by other constants that occur as argument (\small\texttt{A2}\normalsize) of a corresponding literal in the answer set. The 
second rule specifies that if no positive body literal (\small\texttt{pbl/3}\normalsize) for a rule exists, then we can directly 
process the head ({\small\texttt{head/2}\normalsize) of a rule.

The positive body (\small\texttt{pos\_body\_true/3}\normalsize) of the first rule in (28) is true up to some positive 
body literal with respect to a built-in order. If the positive body is true up to the last positive body literal then the 
whole positive body is true. The first rule in (29) deals with this case; the second rule takes care of the first positive 
body literal, and the third rule makes sure that the positive body literals follow the specified order:

\small
\vspace{-0.1cm}
\begin{itemize}
\item[29.]
\begin{verbatim}
pos_body_true(R, A1, A2) :-
  pos_body_true_up_to(R, F, A1, A2), 
  not pbl_not_last(R, F).
\end{verbatim}
\begin{verbatim}  
pos_body_true_up_to(R, F, A1, A2) :- 
  pbl_in_AS(R, F, A1, A2), 
  not pbl_not_first(R, F). 
\end{verbatim}
\begin{verbatim}  
pos_body_true_up_to(R, F1, A1, A2) :- 
  pbl_in_AS(R, F1, A1, A2), 
  F2 < F1, 
  not pbl_in_between(R, F2, F1),
  pos_body_true_up_to(R, F2, A1, A2).
\end{verbatim}
\end{itemize}
\vspace{-0.1cm}
\normalsize

The rule ({\small\texttt{pbl\_in\_AS/4}\normalsize) in (30) checks if a positive body 
literal (\small\texttt{pbl/2}\normalsize) for a rule (\small\texttt{R}\normalsize) exists, looks
in the current answer set ({\small\texttt{in\_AS/1}\normalsize) for a literal
that has the same functor name (\small\texttt{F}\normalsize) as the body literal 
but shows a different argument (\small\texttt{A2}\normalsize)
and returns that argument:

\small
\vspace{-0.1cm}
\begin{itemize}
\item[30.]
\begin{verbatim}
pbl_in_AS(R, F, A1, A2) :-    
  pbl(R, lit(F, A1)), 
  in_AS(lit(F, A2)), 
  A1 != A2.
\end{verbatim}
\end{itemize}
\vspace{-0.1cm}
\normalsize

There exist similar rules that deal with cases where the positive body literal has more than one argument. The
successor relation on positive body literals of each rule is defined with the help of the following auxiliary rules:

\small
\vspace{-0.1cm}
\begin{itemize}
\item[31.]
\begin{verbatim}
pbl_in_between(R, F1, F3) :- 
  pbl(R, lit(F1, A1)), 
  pbl(R, lit(F2, A2)), 
  pbl(R, lit(F3, A3)), 
  F1 < F2, F2 < F3.
\end{verbatim}
\begin{verbatim}  
pbl_not_last(R, F1) :- 
  pbl(R, lit(F1, A1)), 
  pbl(R, lit(F2, A2)), 
  F1 < F2.
\end{verbatim}
\begin{verbatim}  
pbl_not_first(R, F1) :- 
  pbl(R, lit(F1, A1)), 
  pbl(R, lit(F2, A2)),  
  F2 < F1.
\end{verbatim}
\end{itemize}
\vspace{-0.1cm}
\normalsize

The negative part of the body ({\small\texttt{neg\_body\_false/3}\normalsize) in the first rule of (28)
is false, if it can be shown that one of its literals is in the answer set (\small\texttt{in\_AS/1}\normalsize). 
The rule in (32) checks this condition for literals with one argument (other rules deal with literals that have 
more than one argument):

\small
\vspace{-0.1cm}
\begin{itemize}
\item[32.]
\begin{verbatim}
neg_body_false(R, A1, A2) :-
  nbl(R, lit(F, A1)), 
  in_AS(lit(F, A2)).
\end{verbatim}
\end{itemize}
\vspace{-0.1cm}
\normalsize

Finally, the rule \small\texttt{pos\_body\_exists/1} \normalsize in (33) is used as part of the second rule in (28) 
and simply checks if a positive body literal (\small\texttt{pbl/2}\normalsize) exists:

\small
\vspace{-0.1cm}
\begin{itemize}
\item[33.]
\begin{verbatim}
pos_body_exists(R) :- pbl(R, L).
\end{verbatim}
\end{itemize}
\vspace{-0.1cm}
\normalsize

After parsing and translating the CNL text in (5) into reified rules represented as a set of facts, 
the ASP meta-interpreter will generate the following answer set as solution:

\small
\vspace{-0.1cm}
\begin{itemize}
\item[34.]
\begin{verbatim}
{ in_AS(lit(func(student), arg(john))) 
  in_AS(lit(func(work), arg(john)))
  in_AS(lit(func(student), arg(sue)))
  in_AS(lit(func(work), arg(sue))) 
  in_AS(lit(func(student), arg(mary_ann))) 
  in_AS(lit(func(absent), arg(mary_ann)))
  in_AS(lit(func(successful), arg(sue)))
  in_AS(lit(func(successful), arg(john)))
  in_AS(lit(func(neg(work)), arg(mary_ann))) }
\end{verbatim}
\end{itemize}
\vspace{-0.1cm}
\normalsize

\noindent This answer set contains the same information as the answer set in (3) and can be used for question answering.

\section{Conclusion}

In this paper, we investigated in an experimental way if it is possible to process a controlled 
natural language entirely in ASP and if ASP can serve as a unified framework for parsing, 
knowledge representation and automated reasoning. ASP is a powerful declarative knowledge 
representation language that provides support for non-monotonic reasoning and this 
makes the language particularly attractive for controlled natural language processing. We 
showed in detail how a grammar for a controlled natural language can be written as an ASP 
program. This grammar is processed bottom-up and the syntax trees are constructed starting
from the leaves up to the root. The resulting syntax trees are translated into reified rules
that consist of a set of facts. These facts are then used by a meta-interpreter written in ASP
for automated reasoning. The translation into reified rules is necessary because 
ASP does not provide a mechanism that would allow us to generate and assert normal 
ASP rules in the same program. Alternatively, we could take the resulting syntax trees and 
translate them outside of the ASP program into normal ASP rules and then generate a new 
ASP program that executes these rules. With the help of the presented ASP grammar it is
possible to generate look-ahead information to guide the writing process of the author. It
is also possible in ASP to perform anaphora resolution over the reified rules during the translation
process using the standard constraints on anaphoric accessibility. We believe that ASP is an interesting 
paradigm for controlled natural language processing and plan to extend the presented 
approach or aspects of it and integrate them into a controlled language authoring system.


\end{document}